Check for updates

# CTourLLM: enhancing LLMs with Chinese tourism knowledge

Qikai Wei[1,2] · Mingzhi Yang[4] · Jinqiang Wang[1,3] · Wenwei Mao[1,3] · Jiabo Xu[2] · Huansheng Ning[1,3]

## Abstract

Recently, large language models (LLMs) have demonstrated their effectiveness in various natural language processing (NLP) tasks. However, the lack of tourism knowledge limits the performance of LLMs in tourist attraction presentations and travel planning. To address this challenge, we constructed a supervised fine-tuning dataset for the Chinese culture and tourism domain, named Cultour. This dataset consists of three parts: tourism knowledge base data, travelogues data, and tourism QA data. Additionally, we propose CTourLLM, a Qwen-based model supervised fine-tuned with Cultour, to improve the quality of information about attractions and travel planning. To evaluate the performance of CTourLLM, we proposed a human evaluation criterion named RRA (Relevance, Readability, Availability), and employed both automatic and human evaluation. The experimental results demonstrate that CTourLLM outperforms ChatGPT, achieving an improvement of 1.21 in BLEU-1 and 1.54 in Rouge-L, thereby validating the effectiveness of the response outcomes. Our proposed Cultour is accessible at https://github.com/mrweiqk/Cultour.

**Keywords** LLMs · Supervise fine-tuned · NLP · Tourism · Human evaluation

Qikai Wei and Mingzhi Yang have contributed equally to this work.

✉ Huansheng Ning
ninghuansheng@ustb.edu.cn

Qikai Wei
weiqikai@xs.ustb.edu.cn

Mingzhi Yang
asanseu@163.com

Jinqiang Wang
jqwang@xs.ustb.edu.cn

Wenwei Mao
maowenwei@xs.ustb.edu.cn

Jiabo Xu
xujiabo_math@aliyun.com

1 School of Computer and Communication Engineering, University of Science and Technology Beijing, Beijing 100083, China

2 Key Laboratory of Xinjiang Coal Resources Green Mining, Ministry of Education, Xinjiang Institute of Engineering, Urumqi 830023, China

3 Beijing Engineering Research Center for Cyberspace Data Analysis and Applications, Beijing 100083, China

4 Guangxi Tourism Development One-Click Tour Digital Cultural Tourism Industry Co.,Ltd, Guangxi 530012, China

## 1 Introduction

Large language models (LLMs) such as ChatGPT[1][1], ChatGLM[2][2] and Llama[3][3] have proven their effectiveness in various natural language processing (NLP) tasks with their excellent performance [4–7]. These models generate natural and fluent language expressions in diverse contexts by capturing linguistic patterns from vast amounts of textual data, significantly advancing NLP technology in various fields [8–10].

The applications of LLMs in the tourism domain are evident, such as personalized recommendations [11], language translation [12], and chatbots [13]. However, current LLMs have problems with incomplete coverage of cultural and tourism knowledge and insufficient response accuracy. Figure 1 shows an example of a user's conversation with GPT4. GPT4's response to the number of scenic spots is correct, but there is a deviation in the name of the specific scenic spot. This example highlights the urgent need to improve the professionalism of LLMs in tourism. Therefore, the core issue currently faced is how to effectively collect high-quality cultural and tourism data and use the data to train LLMs. Thereby strengthening the knowledge reserve of LLMs in cultural and tourism.

To address these challenges, researchers have explored a range of vertical domain fine-tuning methods to alleviate the lack of corpus in the domain, such as Parameter-Efficient Fine-Tuning(PEFT) [14], Low-Rank Adaptation of Large Language Models(LoRA) [15], and Quantized LoRA(QLoRA) [16]. For instance, researchers have constructed domain-specific datasets in the medicine and law domain, and employed techniques like continued pre-training, vertical domain fine-tuning, and learning from human feedback [17] to enhance model performance. This approach has driven the development of LLMs in specific domains, resulting in more specialized outcomes such as HuaTuo for Chinese medical knowledge [18] and Lawyer LLaMA for legal knowledge [19]. Developing tourism LLMs is crucial for enhancing their performance in providing travel recommendations and improving user experience.

In this paper, we construct a Chinese Supervised Fine-Tuning(SFT) dataset for the culture and tourism domain, named Cultour, which consists of three parts. Firstly, we build a knowledge base of common tourist questions and answers(QA). Leveraging ChatGPT, we design a set of prompts to expand the QA pairs and generate tourism knowledge QA data. Secondly, we organize data related to travel planning to design travelogues manually. Thirdly. We craft unique QA in the aspects of eating, living, traveling, touring, shopping, and entertaining to enrich the diversity of the dataset. Based on this dataset, we propose CTourLLM, a model fine-tuned based on Qwen1.5, to improve the quality of information about attractions and travel planning. To better evaluate the model's performance, we adopt both manual and automatic evaluation methods to assess the performance of CTourLLM. In particular, for manual evaluation, we designed a new metric called RRA (Relevance, Readability, Availability) to evaluate LLMs in the tourism domain manually, and the experimental results demonstrate the effectiveness of CTourLLM.

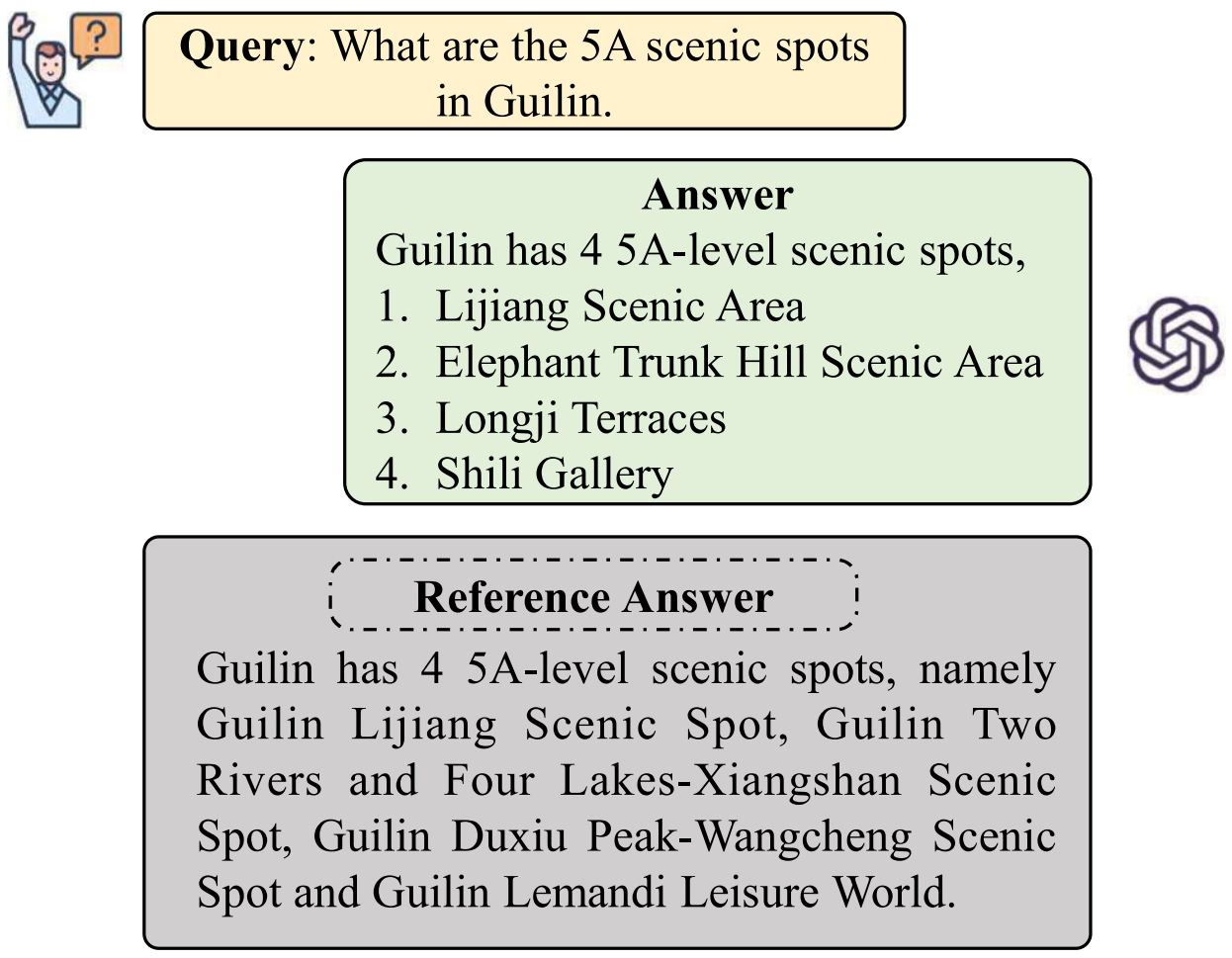

**Fig. 1** Example of a user's conversation with GPT4 on tourism

In summary, our contributions can be summarized as follows:

1. We construct Cultour, a high-quality Chinese SFT dataset for tourism and culture. The dataset contains a tourism knowledge base, travelogues, and tourism QA data.
2. We propose CTourLLM, a Qwen-based model supervised-finetuned with Cultour in the tourism domain.
3. We introduce RRA, a novel metric for evaluating LLMs in the tourism domain, considering relevance, readability, and availability.
4. We assess CTourLLM using both automated and human evaluations. The experimental results demonstrated the effectiveness of CTourLLM.

[1] ChatGPT can be accessed onhttps://openai.com/chatgpt/overview/.

[2] ChatGLM can be accessed on https://chatglm.cn/main/alltoolsdetail?lang=en.

[3] Llama can be accessed on https://www.llama.com/.

## 2 Related work

The emergence of general LLMs such as ChatGPT 3.5 has broken through the limitations of traditional pre-training model methods [20, 21], opened up a new lifestyle for people, and reduced the cost of living [22]. However, in specific application domains, the performance of general large models is still limited [19]. Therefore, researchers are considering combining domain-specific data with general-domain large models to train a vertical-domain large model to serve specific domains. Huang et al. [19] used 50k legal domain-related data to continue pre-train the model, enabling it to possess more relevant knowledge in the legal domain. In answering questions, the model's responses are more knowledgeable. Yang et al. [23] implemented a complete LLMs training process involving pre-training, Supervised Fine-Tuning (SFT), and Reinforcement Learning from Human Feedback (RLHF) to enhance the model's performance in the Chinese medical field. To address the limited datasets for specific languages, Xiong et al. [24] designed a translation model to translate English datasets into Chinese and then fine-tuned the ChatGLM to expand the domain-specific datasets to enhance the model's performance. Li et al. [25] fine-tuned LLaMA using 100K patient-doctor dialogues and supplemented knowledge with a recall-based retrieval system. Wang et al. [18] integrated distilled data from ChatGPT and real data from doctors, leveraging reinforcement learning to enhance model performance and achieve the functionality of interactive medical consultation. Beyond conventional LLMs training methods, Xu et al. [26] employed RLHF [17] and Direct Preference Optimization [27] to make the final output results more fitting to human reading habits. Hallucinations still occur within the domain after optimizing LLMs through continued pre-training, supervised fine-tuning, and RLHF methods. Retrieval-augmented Generation(RAG) [28] combines LLMs (such as Llama and ChatGLM) with external data, enabling the model to refer to the knowledge base when answering questions. Cui et al. [29] used laws and regulations as an external knowledge base to enhance the accuracy of the model's responses. Prompt learning-based methods are also often used to optimize the performance of large models. Zhao et al. [30] proposed the PromptMR method, which uses prompt learning to improve the performance of metaphor resolution, thereby correctly identifying metaphorical expressions in sentences.

The combination of LLMs with the tourism domain has also attracted researchers' attention. Hsu et al. [31] discussed opportunities and challenges of applying Generative Artificial Intelligence (GenAI) and LLMs by the tourism industry and tourists. Mo et al. [32] leveraged prompt-based learning with LLMs, meticulously designing prompts to ensure the responses of the model are reasonable. Wang et al. [33] used domain-oriented LLMs to transform user inquiries into diverse guidance-seeking contexts and facilitate multi-modal interactions. Secchi et al. [34] proposed a method that integrates LLMs with domain-specific knowledge graphs to optimize hotel services. This approach combines domain knowledge graphs with feature engineering to enrich the data representation within LLMs. Based on the current development of the cultural and tourism domain, we constructed the Cultour dataset. Based on this dataset, we proposed CTourLLM, a model fine-tuned based on LLMs to enhance the performance of LLMs in the cultural and tourism domains.

## 3 Our approach

In this section, we provide a detailed overview of our proposed method for constructing the Cultour dataset and the PEFT method for fine-tuning CTourLLM. The overall training process is shown in Fig. 2.

### 3.1 Cultour dataset construction

People appreciate local attractions in tourism and show a strong interest in cultural heritage and intangible cultural assets. However, the training data for general LLMs in the cultural and tourism domain is relatively limited, resulting in suboptimal performance. To address this challenge, we constructed the Cultour dataset to enrich the culture and tourism domain data. This dataset includes tourism knowledge base data, travelogues data, and tourism QA data in the tourism domain. The specific data statistics are shown in Table 1.

#### 3.1.1 Tourism knowledge base data

It is worth noting that there is a large amount of tourism-related data on the internet, which includes misinformation, advertisements, and useless text. To utilize tourism resources more accurately, we manually collected common questions and answers about attractions, food, and intangible cultural heritage from Baidu Baike and online travel websites. For data obtained from the websites, We use regular expressions to process the parsed data and then remove duplicate content. To fine-tune LLMs with tourism data, we modify the data according to the SFT format. The format is as follows: {"instruction": "What is the Baise Uprising Memorial Hall like?", "output": "The Baise Uprising Memorial Hall is themed on comprehensively displaying important figures and important events related to the Baise Uprising"}. Therefore, we designed some prompt templates, such as “Please introduce me to {attraction name}” and concatenated the attraction name corresponding to the data with the prompt template. Finally, we get the SFT data in the tourism field. To ensure the accuracy of the answers, we manually verified each one, ultimately constructing a comprehensive tourism knowledge base. Some examples from the Culture dataset are shown in Table 2.

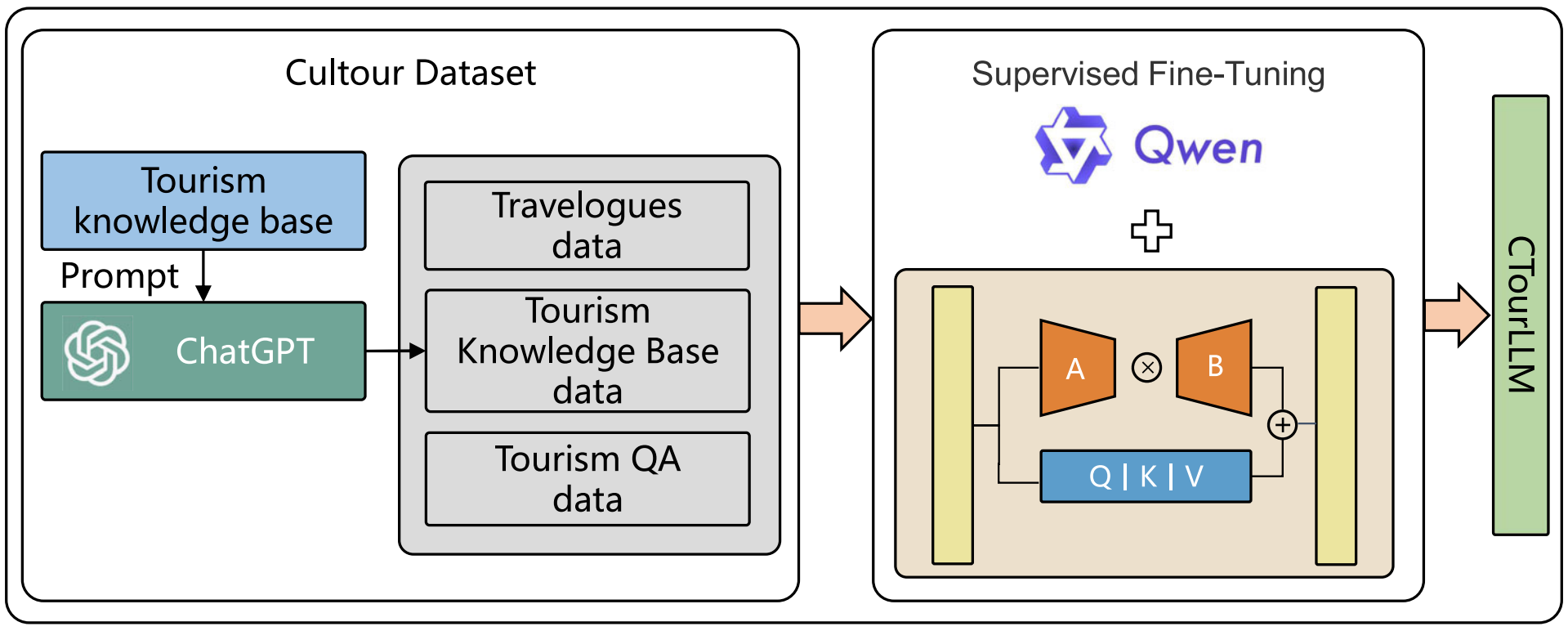

**Fig. 2** The overall training process of CTourLLM

**Table 1** The statistical information of Cultour

| Data type | Number | Average length | Cities involved |
|---|---|---|---|
| Tourism knowledge base data | 9004 | 300 | 14 |
| Travelogues data | 1792 | 1255 | 2 |
| Tourism QA data | 2027 | 159 | 14 |

Inspired by previous research [35], we design a specialized prompt template for generating SFT data based on the structured data in Table 2. The prompt template is shown in Table 3. Utilizing ChatGPT, we convert the knowledge base pairs into single-turn dialogues within the tourism context, enhancing their diversity and richness in language expression. We apply the same processing to all knowledge base data, resulting in 9197 pairs of SFT data in the tourism domain.

In the process of generating answers by ChatGPT, there is a hallucination problem [36], which generates some content that seems reasonable but is incorrect. For example, in the generated results of ChatGPT, there is an error case that describes *“mealworm” as a bug made from insect larvae such as scarab beetles, snake eyes, and bull eyes*. In fact, in this batch of data, *“mealworm”* is a local specialty delicacy whose main ingredient is indica rice. It is named mealworm because it looks like Cordyceps. In this case, ChatGPT’s answer, while logical for the generic domain, is inconsistent with the true answer. To avoid this error, we carefully review the results generated by ChatGPT and manually correct the erroneous parts. As a result, we obtain 9,004 high-quality SFT data points covering attractions, culture, and other tourism domains.

#### 3.1.2 Travelogues data

Travelogues data is open-ended, unlike knowledge base data. We selected data from travel planning provided by some travel agencies and travelogues on tourism platforms such as Ctrip Travel and Little Red Book. We performed manual cleaning, secondary design, and annotation to ensure the data input and output format is consistent with human reading habits. Ultimately, we generated 1792 travelogue SFT data and examples are shown in Table 4.

#### 3.1.3 Tourism QA data

To improve the performance of LLMs in the tourism domain and enhance their understanding of human questions, we expanded the data to meet the versatility of CTourLLM when facing diverse data. It involves many aspects, including eating, living, traveling, touring, shopping, and entertaining. We use ChatGPT to automatically ask questions to collect possible questions and answers, aiming to simulate real user needs, thereby making LLM more accurate and appropriate when answering actual questions. The prompt template we use is “*As a professional travel expert, please generate questions and answers about eating during travel*”. Next, we manually select from the generated questions and answers. Finally, we collected 2,027 QA pairs of SFT data. The data details are shown in Table 5.

### 3.2 Parameter-efficient tuning

To enable the LLMs to understand people’s questions, Wen et al. [37] mentioned that controlling the ratio of domain-

**Table 2** Some examples from the culture dataset

| Question | Reference Answer |
|---|---|
| 桂林漓江景区是几A级景区?<br>What is the rating of Guilin Li River Scenic Area in terms of A-level classification? | 2007年，桂林市漓江景区被评为国家AAAAA级旅游区。<br>In 2007, the Li River Scenic Area in Guilin City was rated as a national AAAAA-level tourist area. |
| 阳朔啤酒鱼是什么?<br>What is Yangshuo beer fish? | 阳朔"啤酒鱼"，是阳朔有名的地方特色菜，鱼肉鲜辣可口，无一丝鱼本身的腥味，啤酒和鱼肉的香味可以让人胃口大开......<br>Yangshuo "Beer Fish" is a famous local specialty dish in Yangshuo. The fish meat is fresh, spicy, and delicious, without any of the fish's inherent fishy taste. The aroma of beer and fish meat can open people's appetites open...... |
| 桂林渔鼓有什么历史渊源?<br>What is the historical origin of Guilin Yu-drum? | 桂林渔鼓是桂林市的汉族传统说唱艺术。明末清初、民国时期都有湖南的渔鼓艺人在桂林传唱渔鼓并落户桂林，开创了桂林渔鼓的先河...<br>Guilin Yu-drum is a traditional Han Chinese narrative and singing art in Guilin City. At the end of the Ming Dynasty and the beginning of the Qing Dynasty, as well as during the Republic of China period, fishermen from Hunan came to Guilin to perform and settle, thus pioneering the tradition of Guilin Yu-drum... |

We translate it into English for better illustration.

specific data to general data at 1:5 enables LLMs to better understand user inquiries and enhances performance in vertical fields [37]. Accordingly, we compiled a training corpus for CTourLLM, consisting of 51K general domain datasets [38] combined with the Cultour dataset. We manually designed a test set of 60 questions covering eating, living, traveling, touring, shopping, and entertaining. Inspired by previous studies [39], fine-tuning the entire model based on given domain-specific SFT data requires many computing resources. Therefore, we use a parameter-efficient fine-tuning method to train LLMs. This method enables us to fine-tune the model with less computing resources. Among these methods, we use LoRA [15] as the fine-tuning method for CTourLLM.

Specifically, the LoRA method freezes the parameters of LLMs and introduces an additional trainable low-rank decomposition matrix in each transformer layer, as shown in Fig. 3. The blue part in Fig. 2 represents the pre-trained weights of frozen LLMs, and only the parameters in the red part are trained. In $y = \mathrm{W}x$, W denotes the pretrained $n \times k$ parameter matrix. We compute $y$ by introducing low-rank matrices A($\mathbb{R}_{n\times r}$) and B($\mathbb{R}_{r\times k}$).

$$y = (\mathrm{W})x + (\Delta\mathrm{W})x = \mathrm{W}x + BAx, B \in \mathbb{R}^{n\times r}, A \in \mathbb{R}^{r\times r}, \tag{1}$$

In Eq. (1), $r$ denotes the rank of A and B, and $r$ is much smaller than min($n$, $k$). Only parameters A and B participate in the model training, so we can complete the training of model parameters at a smaller cost and improve model performance.

## 4 Experimental setup

To validate the performance of the CTourLLM model, we evaluate it against common LLMs and conduct a detailed analysis of the results. Additionally, we introduce RRA, a human evaluation standard specifically for LLMs in the tourism domain, which assesses the model's performance based on Relevance, Readability, and Availability.

**Table 3** The tourism prompt template

| |
|---|
| User prompt: |
| 1. Assume you are a professional AI travel assistant. Based on the provided <question, answer> pairs, generate a logical cultural tourism scenario dialogue. |
| 2. The dialogue should consist of a question or request followed by an answer. The question should be based on the given question and provide complete contextual information. It should be specific and avoid overly technical terms. You can modify the original question to ensure the generated dialogue is logical and fits the scenario. The length should be 1-3 sentences, and the question should have a consulting tone. |
| 3. The answer should be derived from the provided answer, with some enrichment to the content. The answer should address the question in detail, be 15-20 sentences long, and be friendly, approachable, patient, and comprehensive. |
| 4. Enrich the content in the generated response to make it more suitable for human reading habits. |

## 4.1 Baseline

To evaluate the effectiveness of our method, we compare CTourLLM with ChatGPT [17], ChatGLM3 [2], Alpaca [40], Qwen1.5 [41], Ctrip wendao.

**ChatGPT 3.5** is an NLP model designed by OpenAI, which uses manually annotated data for training and reinforcement learning to enhance the model's capabilities.

**ChatGLM3** is an open-source model in the ChatGLM series, characterized by its low deployment barrier. Additionally, it demonstrates strong performance across various datasets in semantics, mathematics, reasoning, code, knowledge, and other domains.

**Alpaca**, derived from LLAMA and continued pretrained with Chinese textual data, exhibits robust proficiency in Chinese comprehension. Furthermore, it uses Chinese instruction data for fine-tuning to enhance the ability of the model to understand the ability of human instructions.

**Qwen1.5**, an encoder-only Transformer model, accommodates prompt with a maximum length of 32K tokens and is compatible with multiple languages, including English, Chinese, French, and Spanish. Additionally, it has been made open-source in various sizes, including 0.5B, 1.8B, 4B, 7B, 14B, and 72B.

**Ctrip wendao** is an LLM designed for tourism, proposed by Ctrip. It can conduct cultural and tourism Q&A, personalized itinerary planning, and real-time travel suggestions. However, Ctrip Wendao is currently limited to use within the Ctrip APP and has not yet opened a public API.

## 4.2 Experimental settings

In this section, we introduce the parameters involved in the fine-tuning CTourLLM process. We set the maximum length of the input sequence to 1024 and 3 epochs. We set the learning rate (Lr) to 5e-4 and employ learning rate warm-up with a setting of 100 [42]. In LoRA, The rank $r$ is set to 8, the constant $a$ is set to 16, and the dropout is set to 0.1. We utilize the Adam[43] optimizer to update the LoRA parameters. All experiments are conducted on 2 Nvidia GeForce RTX 3090 GPUs.The settings of hyperparameters are shown in Table 6.

## 4.3 Evaluation metrics

We evaluate the performance of CTourLLM through three metrics: BLEU [44], Rouge [45], and Meteor [46].

Bilingual Evaluation Understudy(BLEU) [44] is an evaluation metric used to measure the accuracy of models with multiple correct output results. It compares the overlap of n-grams between candidate translations and reference translations. It is commonly used to evaluate machine translation quality.

Recall-Oriented Understudy for Gisting Evaluation(ROUGE) [45] is a set of metrics used to evaluate automatic summarization and machine translation. It measures the "similarity" between an automatically generated abstract or translation and a set of manually generated reference abstracts by calculating the corresponding score.

Metric for Evaluation of Translation with Explicit ORdering(Meteor) [46] is a metric used to evaluate the quality of machine-translation output. Compared to BLEU, METEOR considers more factors such as synonym matching, stem matching, and word order, making it generally considered a more comprehensive evaluation metric.

# 5 Results

In this section, we use automatic and human evaluation methods to evaluate the performance of CTourLLM.

## 5.1 Automatic evaluation

To evaluate the performance of LLMs, we designed 60 questions based on eating, living, traveling, touring,

**Table 4** Travelogues data example

| |
|---|
| Prompt：筹备一份南宁一天的旅游攻略。<br>Prompt: Prepare a ONE-day travelogue to Nanning. |
| Response：接下来的内容将介绍南宁一日游的游记建议，包括主要景点和活动。上午：青秀山：早上可以去青秀山感受阳光和清新空气，步道盘山而上，慢慢将整个城市尽收眼底。下午：广西民族博物馆：紧邻青秀山，是以广西民族文化为专题的博物馆。广西民族博物馆以收藏、研究和展示广西12个世居民族的传统文化为主要工作任务，同时兼顾对广西周边省份各民族以及东南亚各民族的文化研究、文物资料收藏和宣传展示。晚上：西关夜市：步行去西关夜市，小吃种类很多，体验特色美食填饱胃的感觉。以上是南宁一天的旅行指南，希望对您有所裨益。在旅途中，请务必注意安全，保持良好的心境和状态。祝您旅途愉快！<br>Response: The following content will introduce travelogues suggestions for a day trip to Nanning, including the main attractions and activities. Morning: Qingxiu Mountain: In the morning, you can go to Qingxiu Mountain to feel the sunshine and fresh air. The trail winds up the mountain and slowly takes in the entire city. Afternoon: Guangxi Museum of Ethnology: Close to Qingxiu Mountain, it is a museum with the theme of Guangxi ethnic culture. The main mission of the Guangxi Ethnology Museum is to collect, research, and display the traditional culture of the 12 living ethnic groups in Guangxi. At the same time, it also takes into account the cultural research, collection of cultural relics and publicity and display of various ethnic groups in the surrounding provinces of Guangxi. Evening: Xiguan Night Market: You can walk to Xiguan Night Market. There are many kinds of snacks, and you can experience the feeling of filling your stomach with special delicacies. The above is a one-day travelogues guide in Nanning, I hope it will be helpful to you. During the journey, please pay attention to safety and maintain a good mood and condition. I wish you a pleasant journey! |

shopping, and entertaining. We collected all the responses without any prompt templates from LLMs. For Responses, we use BLEU-1 (B-1), BLEU-2 (B-2), BLEU-3 (B-3), ROUGE-1 (R-1), ROUGE-2 (R-2), ROUGE-L (R-L), and Meteor as the metrics to evaluate LLMs, the result is shown in Table 7.

In Table 7, the CTourLLM model achieved optimal performance on most metrics. On the BLEU-related metric, the performance of CTourLLM is much higher than other models. On the Rouge-L metric, CTourLLM achieved the second-best performance. On the meteor metric, CTourLLM achieved optimal performance compared with other models. Experimental Results prove that the CTourLLM model can improve QA effects in the culture and tourism domains.

## 5.2 Human evaluation

Inspired by the previous literature [47], we conduct the human evaluation to assess the acceptability of the results generated by LLMs. We designed sixty questions for LLMs and collected responses from ChatGLM3, Qwen1.5, Llama-Chinese, CTourLLM, and ChatGPT 3.5. To facilitate a fair evaluation, we devise a set of human evaluation metrics: Relevance (the degree of understanding queries and providing relevant answers), Readability (correctness of grammar, fluency, and formatting level), and Availability (reflects the understanding of the content of tourist attractions, correctness of the responses, such as information about transportation to attractions, relevant introductions). Each metric is scored on a scale of 0 to 3, the specific scoring criteria outlined in Table 8.

We manually designed a test set of 60 questions covering eating, living, traveling, touring, shopping, and entertaining. We used five models to answer the 60 questions, respectively, and the testers were required to rate each answer in terms of relevance, readability, and availability. We invited a total of 10 travel enthusiasts to participate in the test. Each participant spent at least 30 min completing the test without any interruptions. In the evaluation process, we randomize the order of queries, and the final score is the average of all the scores, as shown in Table 9. The results show that our CTourLLM significantly improves the availability of knowledge in the culture and tourism domains without compromising model readability.

## 5.3 Ablation experiment

To investigate the impact of various components of the Cultour dataset on our experimental results, we conducted a series of ablation studies. We followed the methodology outlined in Ref. [39], maintaining a 1:5 ratio between domain-specific and general data, with a learning rate of 5e-4, to evaluate the model's performance. Since the

**Table 5** Some examples from the tourism QA SFT data

| Question | Reference Answer |
| --- | --- |
| 考虑到在桂林旅游的过程中有小孩（老人或其他特殊人群），帮我推荐一下适合跟他们一起的景点?<br>Considering that we will be traveling with children (elderly people or other special groups) in Guilin, could you recommend some attractions suitable for traveling with them? | 当然，考虑到有小孩、老人或其他特殊人群同行，推荐选择一些适合全家游玩的景点。以下是两个推荐：桂林两江四湖：这是一个非常适合全家游玩的景点...<br>Of course, considering that there are children, the elderly or other special groups traveling with you, it is recommended to choose some attractions suitable for the whole family. Here are two recommendations: Guilin Two Rivers and Four Lakes: This is a very suitable attraction for the whole family... |
| 当地的地理环境对广西当地食材的选择和菜肴特点有何影响?<br>What influence does the local geographical environment have on the selection of local ingredients and the characteristics of dishes in Guangxi? | 广西地理环境的多样性和复杂性对当地的食材选择和菜肴特点产生了重要影响。以下是一些例子：1. 气候和地形：广西地处亚热带和热带气候区...<br>The diversity and complexity of Guangxi's geographical environment has a significant impact on the local food selection and cuisine characteristics. Here are some examples: 1. Climate and topography: Guangxi is located in the subtropical and tropical climate zone... |

We translate it into English for better illustration

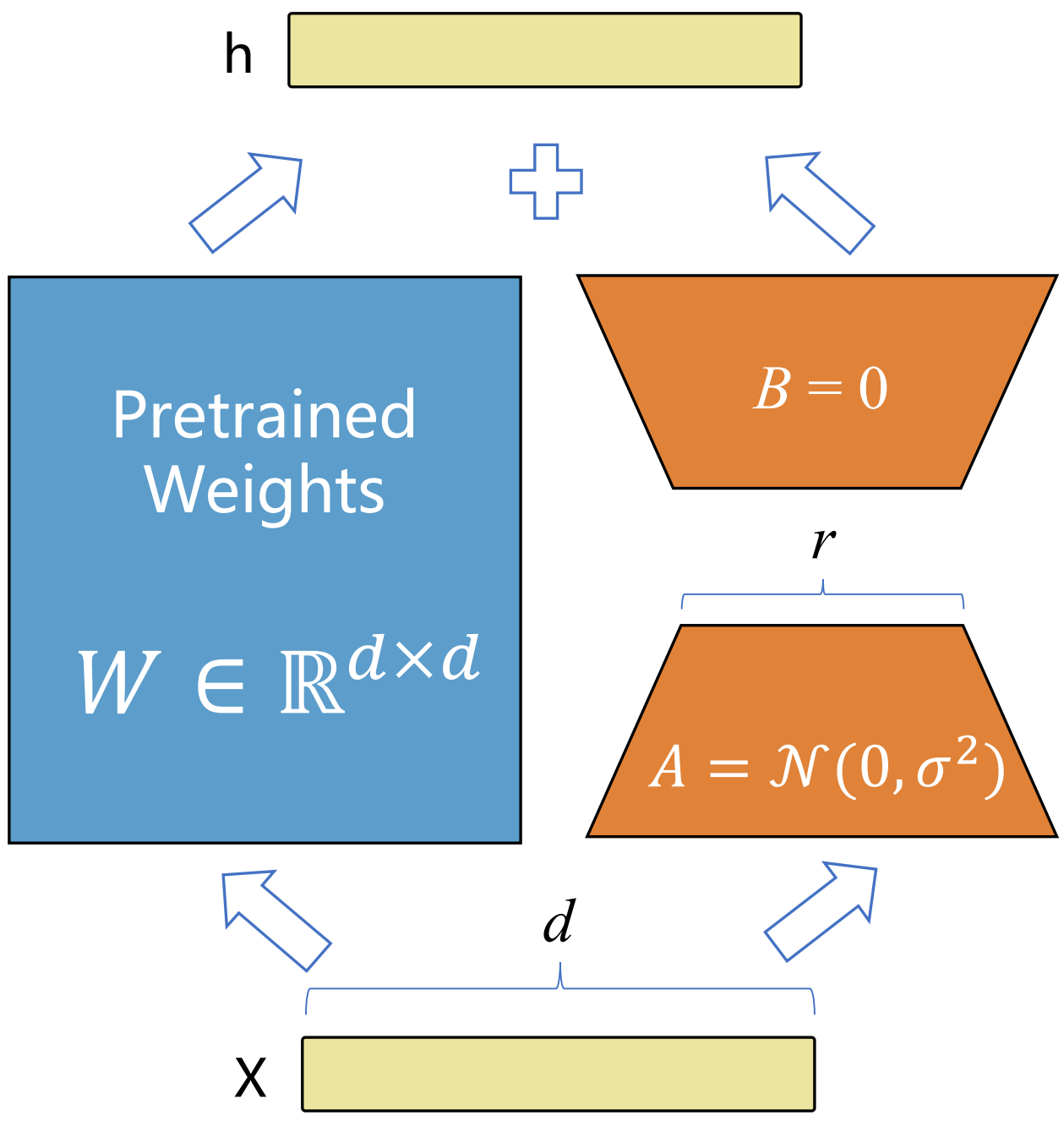

**Fig. 3** Illustration of LoRA method

Tourism KB dataset is large, we divided it into 2000 examples and conducted detailed ablation experiments. The results of these experiments are presented in Table 10.

**Table 6** The hyperparameters used in the training process

| Hyperparameters | Settings |
| --- | --- |
| Learning rate | 5e-4 |
| Maximum length | 1024 |
| Number of epochs | 3.0 |
| Rank | 8 |
| Constant | 16 |
| GPU types | Nvidia GeForce RTX 3090 |
| Training time | 535 min |

From the results in the Table 10, we can see that different training corpora have different impacts on model performance, and Tourism KB+Travelogues data have the greatest impact on the model results.

In addition, we found that the model struggled to provide accurate travel plans without fine-tuning with

**Table 7** Experimental results of all models on the Cultour dataset

| Model | B-1 | B-2 | B-3 | R-1 | R-2 | R-L | Met. |
|---|---|---|---|---|---|---|---|
| ChatGPT 3.5 | <u>20.46</u> | <u>9.93</u> | <u>5.04</u> | <u>25.68</u> | <u>8.55</u> | 17.23 | <u>17.96</u> |
| Qwen-1.5-7B | 18.76 | 8.21 | 3.91 | **26.97** | 6.92 | **19.45** | 17.30 |
| ChatGLM3-6B | 19.44 | 8.99 | 4.75 | 24.22 | 7.61 | 18.38 | 17.01 |
| LLama2-chinese-7B | 7.95 | 3.77 | 1.97 | 14.48 | 3.84 | 11.99 | 9.37 |
| Ctrip wendao | 12.54 | 6.00 | 3.21 | 22.38 | 7.17 | 17.61 | 14.20 |
| CTourLLM-7B(Ours) | **21.67** | **12.05** | **7.63** | 25.46 | **10.72** | <u>18.77</u> | **18.62** |

Bold represents the best result and underlined represents the second-best result travelogues data or generated incorrect information when processing travel plan queries. However, after fine-tuning with travelogues data, the model's performance improved significantly, and the response aligned more with the expected format. This result indicates that SFT data fine-tuning can enhance the model's accuracy in specific tasks, making it more suited to domain-specific requirements. Nonetheless, this fine-tuning might diminish the model's overall general performance. Therefore, when employing SFT data for fine-tuning, it is crucial to balance domain-specific and general performance.

**Table 8** Human evaluation scoring criteria and explanations

| Category | Score | Explanation |
|---|---|---|
| Relevance | 0 | The response is entirely irrelevant to the query. |
| | 1 | The response is only somewhat relevant to the query. |
| | 2 | The response can match the core of the query, but further queries are needed to supplement the answer. |
| | 3 | The response accurately matches the core of the query. |
| Readability | 0 | The answer is extremely difficult to understand or has poor grammar and structure. |
| | 1 | The response is somewhat difficult to understand or has grammatical errors. |
| | 2 | The response is clear, with only a few grammatical errors or room for improvement. |
| | 3 | The response is very clear, well structured, has no grammatical errors, and is easy to understand. |
| Availability | 0 | There are serious problems or errors in the response and the answer is not very usable. |
| | 1 | There are some errors in the response, some of which need to be verified by searching for the response. |
| | 2 | The response is error-free, with only a few confusing elements. |
| | 3 | The response is error-free and instructive, providing a better experience for travelers. |

**Table 9** The results of ablation experiment

| Model | Relevance | Readability | Availability |
|---|---|---|---|
| ChatGPT 3.5 | **2.68** | 2.68 | 2.20 |
| LLama2-chinese-7B | 2.24 | 2.34 | 1.79 |
| ChatGLM3-6B | 2.47 | 2.63 | 2.06 |
| Qwen-1.5-7B | 2.57 | **2.69** | 2.15 |
| CTourLLM-7B(Ours) | 2.62 | 2.65 | **2.29** |

Bold represents the best result

# 6 Discussion

## 6.1 The influence of learning rate

During our evaluation of model performance, we observed that the learning rate significantly impacts the overall performance, especially for vertical domain data. To better understand this effect, we analyzed various learning rates and assessed the model's performance using BLEU-1, BLEU-2, Rouge-1, Rouge-2, Rouge-L, and Meteor metrics. As shown in Fig. 4, the results indicate that the learning rate 5e-4 yields the best performance. Conversely, at the learning rate 1e-3, the model fails to learn effectively, leading to a marked decline in performance. Similarly, when the learning rate is reduced to 1e-5, we observe a degradation in performance. These findings indicate that excessively high and low learning rates can negatively impact model performance.

**Table 10** The impact of different SFT data compositions on the model

| Model | R-1 | R-2 | R-L | Met. |
|---|---|---|---|---|
| CTourLLM | – | – | – | – |
| +Tourism KB_2000 | 17.40 | 3.46 | 11.09 | 11.29 |
| +Tourism KB_4000 | 17.53 | 3.47 | 11.47 | 11.87 |
| +Tourism KB_6000 | 19.70 | 4.18 | 12.36 | 12.30 |
| +Tourism KB_8000 | 20.20 | 4.71 | 12.70 | 12.76 |
| +Tourism KB_all | 19.17 | 5.98 | 13.02 | 12.67 |
| +Travelogues | 15.42 | 2.38 | 11.00 | 10.49 |
| +Tourism QA | 16.53 | 4.63 | 12.47 | 10.10 |
| +Tourism KB+Travelogues | 21.89 | 6.41 | 15.28 | 14.95 |
| +Tourism KB+Tourism QA | 21.11 | 6.36 | 14.61 | 13.50 |
| +Travelogues+Tourism QA | 17.27 | 5.47 | 12.47 | 11.89 |
| +All | **25.46** | **10.72** | **18.77** | **18.62** |

Bold represents the best result

### 6.2 The impact of data scale

We investigate the effect of varying the amount of SFT data on model performance. Specifically, we use 1K, 5K, 10K, 30K, and 60K data to train Qwen 1.5 and utilize BLEU-1, BLEU-2, Rouge-1, Rouge-2, Rouge-L, and Meteor to evaluate performance, and the results are shown in Fig. 5. The model's performance will improve with the improvement of the number of SFT data. Interestingly, when the amount of data is 30K, the model's performance has decreased significantly. One possible reason for this inconsistency is that the data is imbalanced. The model may favor predicting the more frequent categories, leading to decreased prediction performance for the less frequent categories.

## 7 Conclusion

In this paper, we present Cultour, a high-quality Chinese tourism SFT dataset specifically fine-tuned for LLM. It includes 12,823 data items such as eating, living, traveling, touring, shopping, and entertaining. Based on this dataset, we fine-tuned an LLM, CTourLLM, for the cultural and tourism domain to gain deeper insights into people's needs. To better evaluate the performance of CTourLLM, we propose human evaluation RRA metrics tailored for LLMs, incorporating relevance, readability, and availability. We use both automated and manual evaluation methods to assess model performance. Experimental results demonstrate the effectiveness of CTourLLM.

In future work, we will expand the data scale and types based on the existing Cultour dataset, including regional tourism characteristics, cultural background introduction, and user-personalized demand analysis. These data will enable the model to understand better and meet the diverse needs of different user groups. At the same time, we will use CTourLLM and related external documents to enhance retrieval capabilities and improve response reliability.

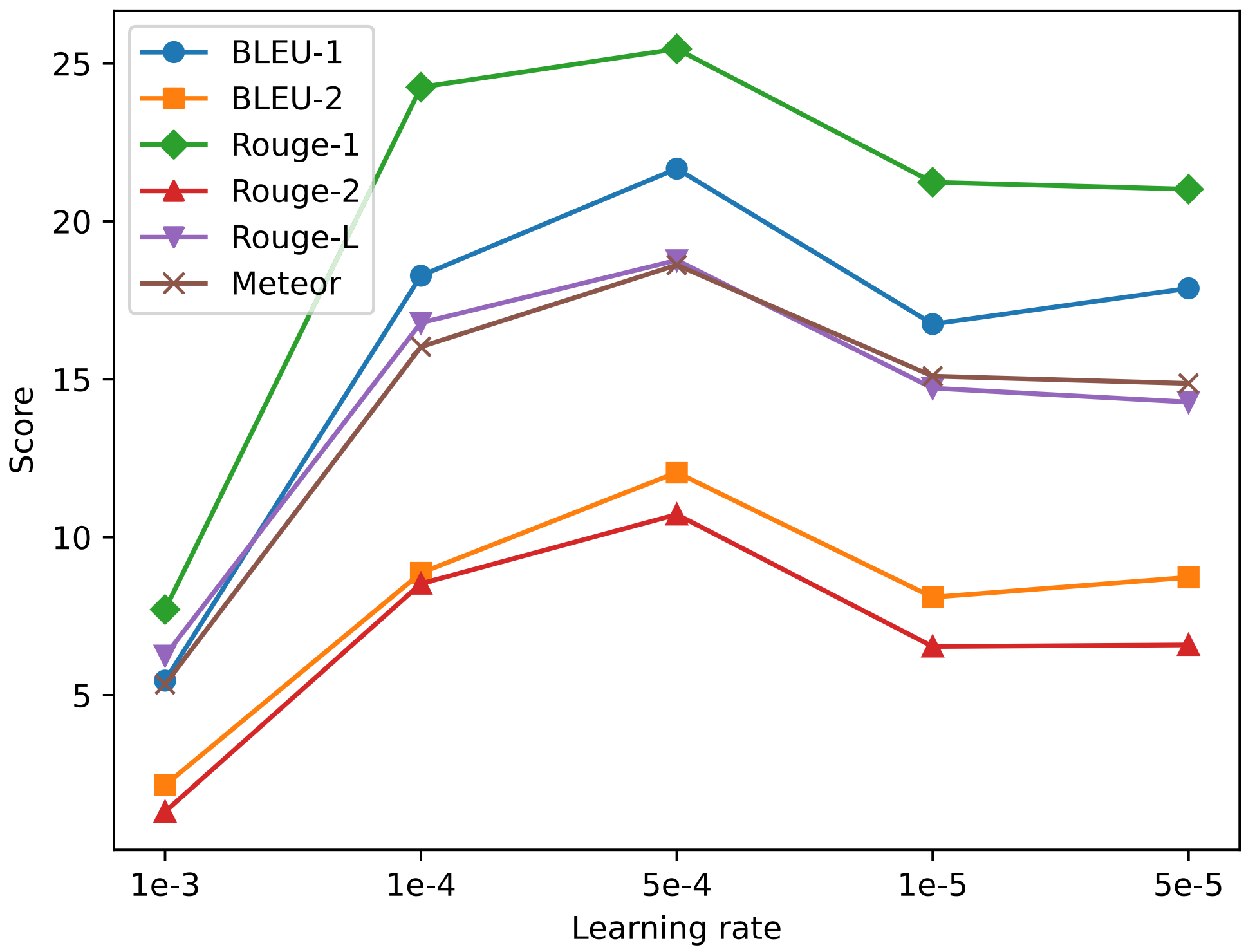

**Fig. 4** The impact of different learning rates on model performance

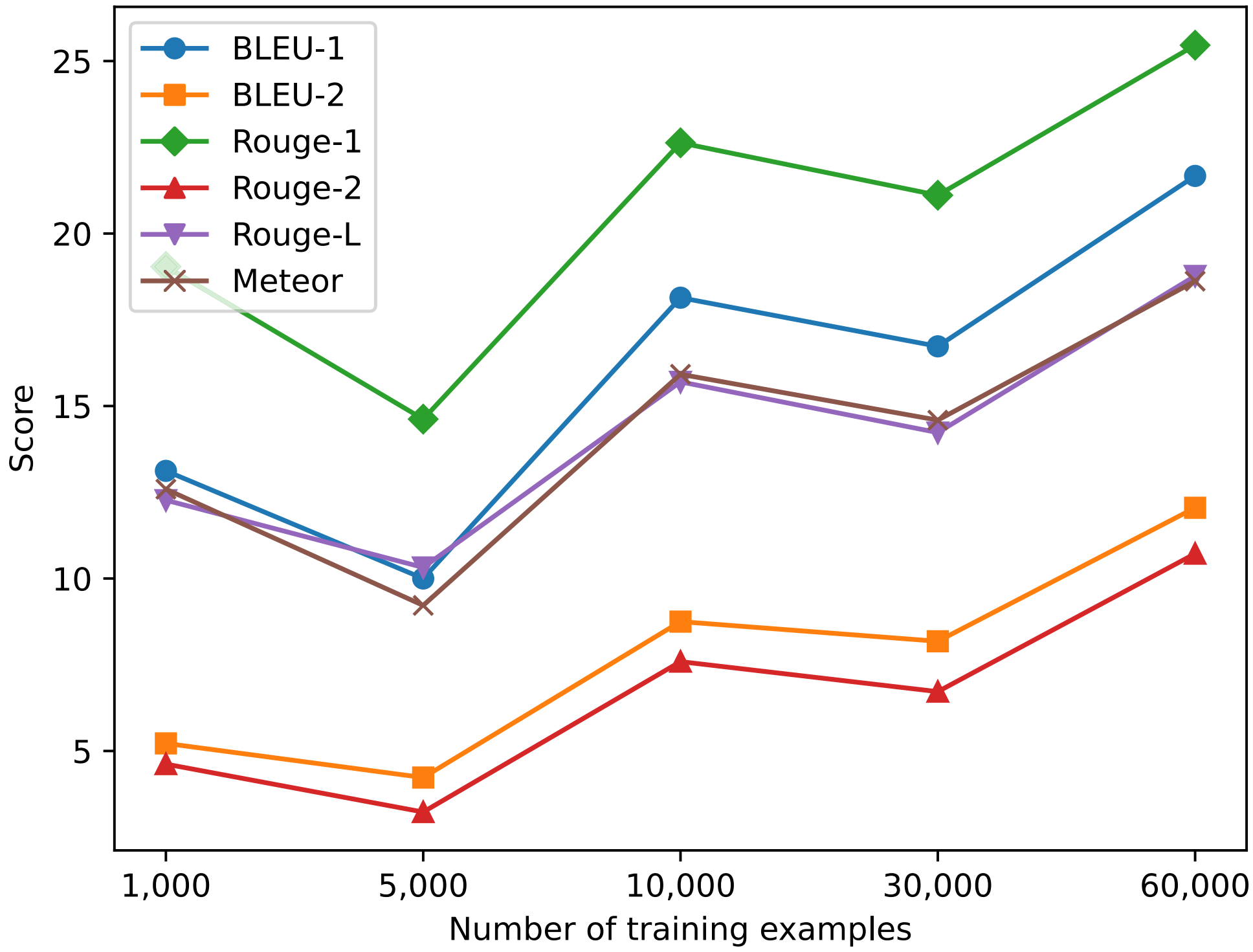

**Fig. 5** The impact of different scales data on model performance

**Author contributions** Q.W., J.W. and W.M. wrote the main manuscript text and Q.W. prepared Figs. 1–4. M.Y., J.X. and H.N.polished the manuscript and checked for grammatical errors. All authors reviewed the manuscript.

**Funding** The authors have not disclosed any funding.

**Data availability** We provide the URL of the data in the abstract of the manuscript.

## Declarations

**Competing interest** The authors declare no competing interests.

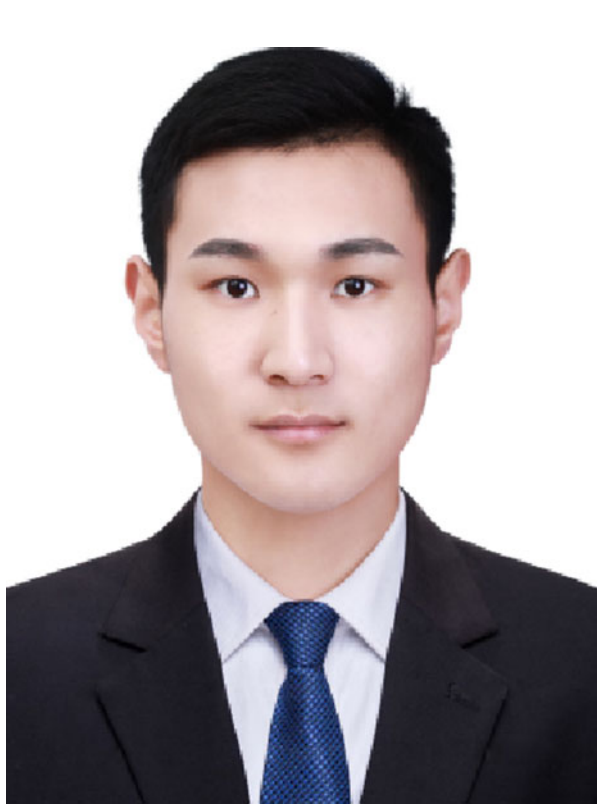

**Qikai Wei** received the B.S. degree from the School of Information Engineering, Hebei University of Geosciences, Shijiazhuang, China, in 2019, and the M.S. degree from the School of Information Science and Engineering, Yanshan University, Qinhuangdao, China, in 2022. He is currently pursuing the Ph.D. degree in the School of Computer and Communication Engineering, University of Science and Technology Beijing. His current research interests include natural language processing and LLMs.

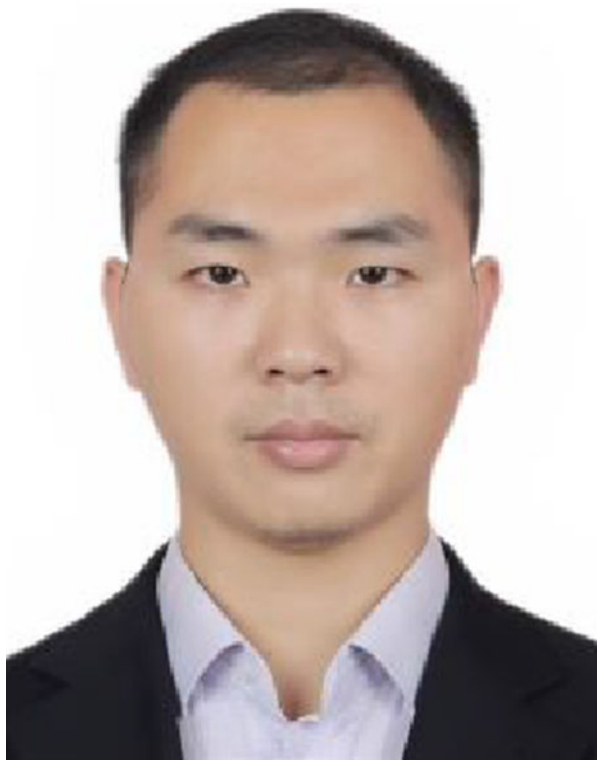

**Mingzhi Yang** received his B.S. degree in School of Mechanical Engineering Southeast University in 2007. He is a senior engineer at Guangxi Tourism Development One-key Tour Digital Cultural Tourism Industry Co., Ltd. His current research direction is large language model and natural language processing.

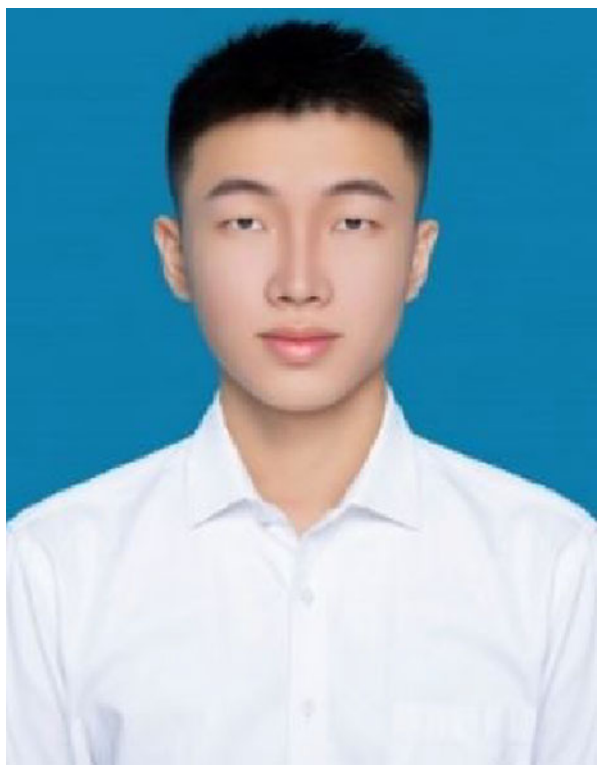

**Jinqiang Wang** received his B.E. degree from Henan Normal University in 2020 and M.E. degree from the University of South China in 2023. He is currently pursuing his Ph.D. degree at the University of Science and Technology Beijing. His research interests include intelligent health, pattern recognition, and natural language processing.

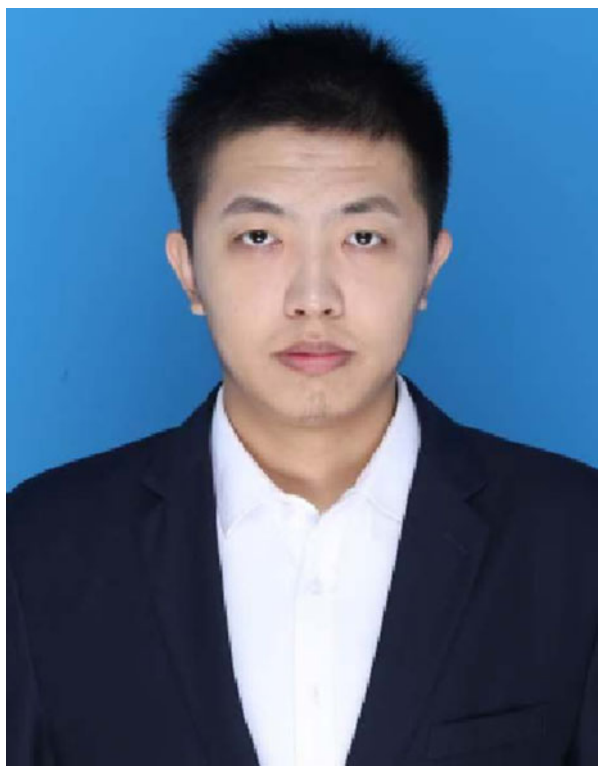

**Wenwei Mao** received the B.S. degree from Wuhan Polytechnic University, Wuhan, China, in 2018, and the M.S. degree from Fujian University of Technology, Fuzhou, China, in 2023. He is currently pursuing the Ph.D. degree in the School of Computer and Communication Engineering, University of Science and Technology Beijing. His current research interests include computer vision and LLMs.

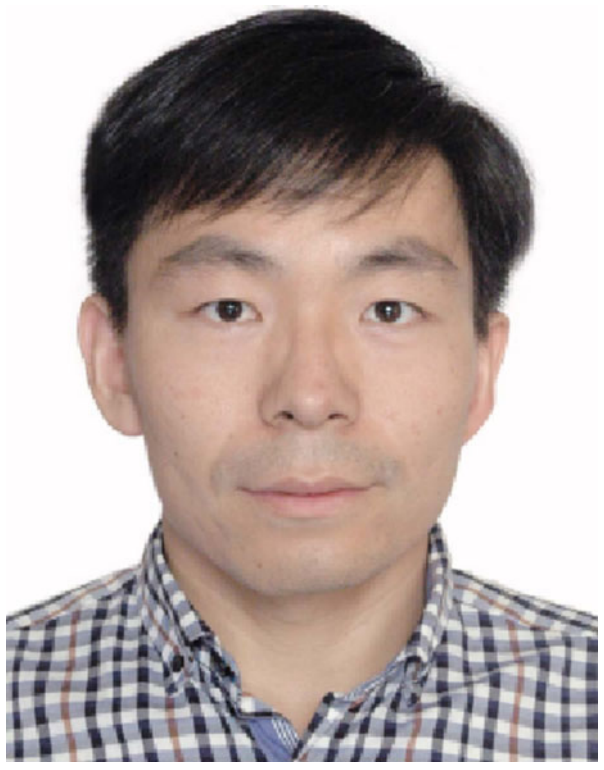

**Jiabo Xu** received the BS degree from Yantai University, China in 2004, and the PhD degree from Xinjiang University, Urumqi, China in 2011. He is currently a professor and the dean of the School of Information Engineering, Xinjiang Institute of Engineering.

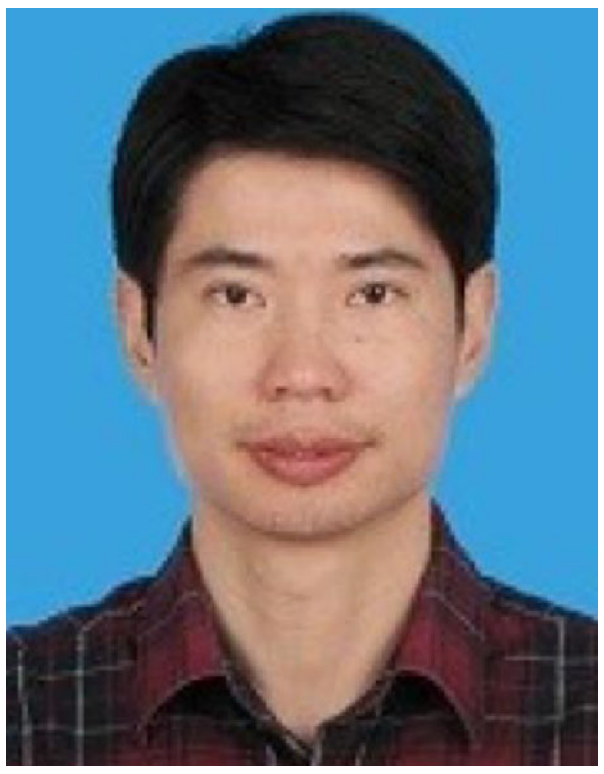

**Huansheng Ning** (IET Fellow) received the B.S. degree from Anhui University, Hefei, China, in 1996, and the Ph.D. degree from Beihang University, Beijing, China, in 2001. He is currently a Professor and the Vice Dean with the School of Computer and Communication Engineering, University of Science and Technology Beijing, Beijing, and Beijing Engineering Research Center for Cyberspace Data Analysis and Applications, Beijing, and the Founder and the Principal of Cybermatics and Cyberspace International Science and Technology Cooperation Base, Beijing. He has authored several books and over 200 papers in journals and at international conferences/workshops. His current research interests include Internet of Things, cyber-physical-social systems. Prof. Ning was a recipient of the awards, including the IEEE Computer Society Meritorious Service Award and the IEEE Computer Society Golden Core Member Award. He has been the Associate Editor of IEEE SYSTEMS JOURNAL and IEEE INTERNET OF THINGS JOURNAL. He was the Chairman in 2012 of the program committee at the IEEE International Internet of Things Conference and the 2015 Smart World Congress. In 2018, he was elected as a Fellow of the Institution of Engineering and Technology. Since 2020, he has been selected into the list of Elsevier's highly cited authors in the field of computer science and Stanford University's top 2% scientists in the world.